\documentclass{mva_style}
\usepackage[pdftex]{graphicx}

\finalcopy %Uncomment this line for the Camera-Ready Manuscript

\begin{document}
\title{Lifelong Change Detection: Continuous Domain Adaptation for Small Object Change Detection in Every Robot Navigation}

\author{
 Koji Takeda\\
 Tokyo Metropolitan Industrial \\
 Technology Research Institute\\
%  2-4-10, Aomi, Koto-ku, Tokyo, Japan\\
 {\tt takeda.koji\_1@iri-tokyo.jp}\\
 \and
 Kanji Tanaka\\
 University of Fukui\\
%  3-9-1 Bunkyo, Fukui-shi, Fukui, Japan\\
 {\tt tnkknj@u-fukui.ac.jp}\\
 \and
 Yoshimasa Nakamura\\
 Tokyo Metropolitan Industrial\\
  Technology Research Institute\\
%  2-4-10, Aomi, Koto-ku, Tokyo, Japan\\
 {\tt nakamura.yoshimasa@iri-tokyo.jp}\\
}

% \author{
% \\
% \\
% \\
% % \\
% }

% }
% \vspace{100mm}

\maketitle

\section*{\centering Abstract}

\textit{
  The recently emerging research area in robotics, ground view change detection, suffers from its ill-posed-ness because of visual uncertainty combined with complex nonlinear perspective projection. To regularize the ill-posed-ness, the commonly applied supervised learning methods (e.g., CSCD-Net) rely on manually annotated high-quality object-class-specific priors. In this work, we consider general application domains where no manual annotation is available and present a fully self-supervised approach. The present approach adopts the powerful and versatile idea that object changes detected during everyday robot navigation can be reused as additional priors to improve future change detection tasks. Furthermore, a robustified framework is implemented and verified experimentally in a new challenging practical application scenario: ground-view small object change detection. 
}

\section{Introduction}
In everyday indoor navigation, robots often need to detect nondistinctive, small change objects (e.g., stationery, lost items, and junk) to maintain their domain knowledge. In the field of machine vision, this is most relevant to the emerging research area of ground-view change detection \cite{chen2021dr,sakurada2020weakly}, which aims to train a change detector in the target domain such that it can successfully detect the change objects from a single ground-view onboard camera without relying on typical change-detection devices (e.g., laser range finders, tactile sensors) or simplified vision setups (e.g., satellite-view parallel projection). This setup is highly ill-posed because of visual uncertainty combined with complex nonlinear perspective projection and remains largely unsolved. Current methods rely on manually annotated high-quality class-specific priors for objects to regularize a change detector model (e.g., CSCD-Net \cite{sakurada2020weakly}). Thus, they are not directly applicable to nondistinctive, small change objects in robotic applications.

\begin{figure}[t]

    \begin{center}
      \includegraphics[height=45mm]{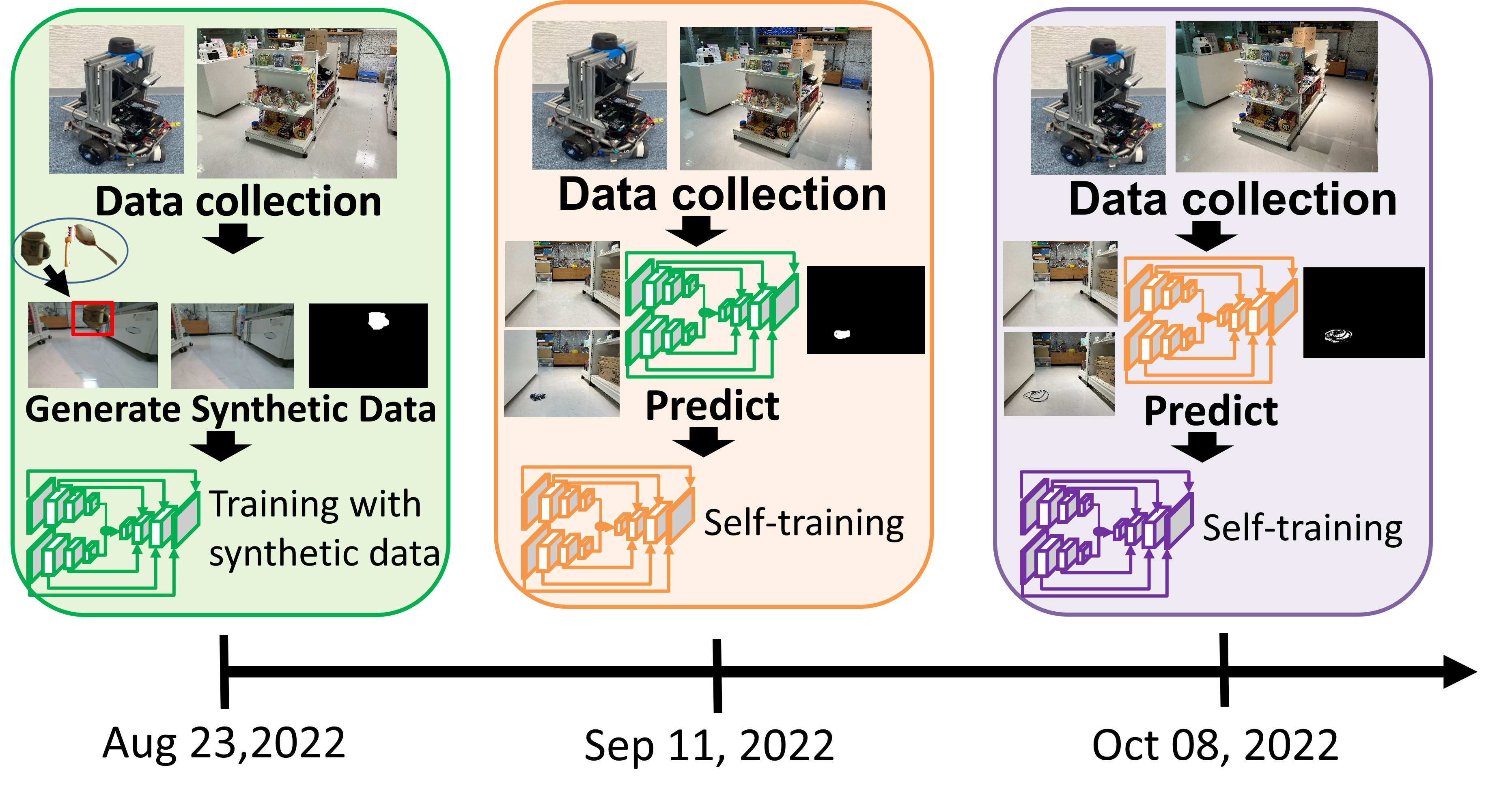}
    \end{center}
    \vspace{-4mm}
    \caption{
      Lifelong change detection scheme. Initialize the domain knowledge with synthetic data and continuously update the domain knowledge by the robot itself.
    }
    \label{tobirae}
    \vspace{-5mm}
  \end{figure}

Motivated by these challenges, we propose a continual domain adaptation (CDA) scheme named Lifelong Change Detection for a new challenging scenario: ground-view small object change detection (GV-SCD) as shown in Figure. \ref{tobirae}. The present approach adopts the powerful and versatile idea that changes in objects detected during everyday robot navigation can be reused as additional priors to improve future change-detection tasks. Every time a change-detection signal arrives, a prior-update module combined with a deep change-detection network is triggered to provide the next-generation updated object knowledge base for data augmentation for improved learning. A difficulty arises from the fact that such an object knowledge base is frequently contaminated by noise and is clearly of lower quality than a manually annotated knowledge base. The proposed scheme addresses this problem in a fully self-supervised manner. Specifically, it begins with unrealistic object instances derived from the COCO dataset \cite{https://doi.org/10.48550/arxiv.1405.0312}. It then continually incorporates noisy and low-quality foreground and background priors collected in the real world such that the change-detection model adapts incrementally to the target domain. The proposed framework is implemented on a real-world indoor mobile robot equipped with a camera and verified experimentally in multiple workspaces.

\section{Related Work}
\subsection{Image Change Detection}

Image change detection is the problem of detecting changes between two scenes and is used for anomaly detection in fixed surveillance cameras \cite{7795955}, farmland analysis from satellite images \cite{9298743}, etc. 
Common problem settings for change detection are satellite imagery\cite{8444434,8618401}.
However, such a simple top-down vision setup often relies on the assumption that pixel-to-pixel correspondences are simplified. In contrast, for ground-view change detection, the problem is highly ill-posed because of complex perspective projection mapping.

For ground-view change detection, a Siamese-based CNN is commonly used. Sakurada et al. verified the effectiveness of CNNs in change-detection applications \cite{jst2015change}. Subsequently, a method using optical flow for image registration was proposed; however, it had a problem with computational cost \cite{sakurada2017dense}. Therefore, a CSCD-Net\cite{sakurada2020weakly} using a correlation layer\cite{dosovitskiy2015flownet} was proposed, and its effectiveness was demonstrated on the public PCD datasets\cite{jst2015change}. Furthermore, the DR-TA Net has been proposed as a method for reducing the calculation cost \cite{chen2021dr}.
However, most existing methods target major object categories such as ImageNet\cite{deng2009imagenet}.
To regularize highly ill-posed ground-view change detection, a fine-grained pixel-level change mask was required as training data.

\subsection{Self Training}
Self-training schemes are commonly used in semi-supervised learning (SSL) areas \cite{amini2022self}. Recently, the scope of self-training has been expanded by several authors to unsupervised domain adaptation\cite{saito2017asymmetric, zou2018unsupervised, zou2019confidence}, where the goal is to transfer knowledge from the labeled source domain to the target unlabeled one. Here, we extend this line of work to investigate the application of change detection.

% Self-training schemes are commonly used in semi-supervised learning (SSL) areas \cite{amini2022self}. Recently, the self-training is not only restricted to semi-supervised learning, but has also been extensively applied for unsupervised domain adaptation \cite{saito2017asymmetric, zou2018unsupervised, zou2019confidence}, where the goal is to transfer knowledge from the labeled source domain to the target unlabeled one. 
% While most of the existing research focus on predicted data, we focus on aligning the distribution of supervised data.

\subsection{Continuous Domain Adaptation}
Unlike standard domain adaptation which assumes a specific target domain, continuous domain adaptation considers the adaptation problem with continually changing target data. 
Continuous Manifold Adaptation (CMA) \cite{hoffman2014continuous} is an early study which considers adaptation to evolving domains. 
Incremental adversarial domain adaptation (IADA) \cite{wulfmeier2018incremental} adapts to continually changing domains by adversarially aligning source and target features. \cite{volpi2021continual} aims to continually adapt to the unseen visual domain while alleviate the forgetting in the seen domain without retaining the source training data. \cite{bobu2018adapting} attempted to adapt to gradually changing domains by assuming continuity between them. 

Ground-view small object change detection has not yet been explored, despite its importance, mainly because domain-specific change masks are not available. In this study, the continuous domain adaptation framework addresses this issue by automatically collecting change mask annotations using the robot itself.

\section{Approach}
The task of continuous domain adaptation (CDA) was initialized in a new workspace, with a single visual experience of a robot navigating the workspace only once as the sole domain knowledge. Two key difficulties differentiate this work from existing work on image change detection. First, the objects to be detected lack semantic features; thus, semantic cues such as semantic relationships \cite{zhu2017learning} or co-occurrence relationships between objects \cite{shih2017deep} and places are not available. Second, a supervised training set, such as training images with ground-truth annotations of region masks of change objects, is not available for robotic applications.

\begin{figure}[t]

  \begin{center}
    \includegraphics[height=20mm]{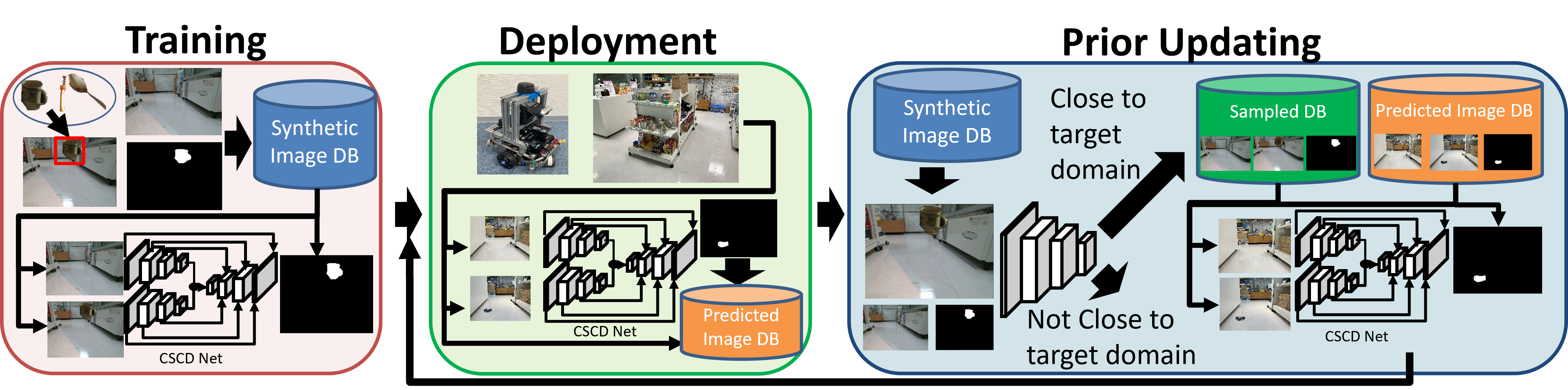}
  \end{center}
  \caption{
    Proposed Lifelong change detection framework.
  }
  \label{proposed}
  \vspace{-5mm}
\end{figure}

The proposed CDA framework comprises three iterative steps: training, deployment, and prior updating (Fig. \ref{proposed}). The first-generation change detector is initialized with a minimal knowledge base consisting of a publicly available COCO object dataset\cite{https://doi.org/10.48550/arxiv.1405.0312} as the sole object prior (Section 3.1). The COCO objects are not in the domain. Nevertheless, as shown in the experimental section, even such a minimal knowledge base is effective and often contributes to improving GV-SCD performance. From the second generation onward, the object knowledge base is incrementally augmented by incorporating the change-detection results (Section 3.2). Moreover, real-world object priors are reused as cues to filter unrealistic objects from the COCO dataset (see Section 3.3). Every time the knowledge base of a new-generation arrives, a deep Siamese network is trained as an image change detector using the knowledge base as the training set (Section 3.4).

\subsection{First Generation Domain Knowledge}
The COCO dataset \cite{https://doi.org/10.48550/arxiv.1405.0312} was chosen as the source of the generic knowledge base for the object priors. The COCO dataset provides a diverse sample of frequently occurring objects in indoor scenes (e.g., stationery, lost items, and junk) with ground-truth annotations. Specifically, training samples were generated using the following procedure: First, one of the images experienced in the previous navigation was sampled and used as a training background image. Next, the COCO object was sampled and pasted to a random size at a random position on the training background image. The resulting synthetic image was used as the training sample.

\subsection{Second Generation and Beyond}
In subsequent navigation, the images detected as having change objects were used as additional training samples. This training sample is domain-specific in two respects. First, it contains knowledge of an object's appearance under realistic lighting conditions, in contrast to the unrealistic appearance in the COCO dataset. Second, it provides domain knowledge regarding the location likelihood of small objects in a certain background scene, in contrast to COCO object images that lack spatial information.

In the proposed scheme, an assumption was made about the smallest detectable object size being a minimum of one pixel. All the detected scene images that satisfied this assumption were incrementally added to the knowledge base.

\vspace{-1mm}

\subsection{Filtering Object Priors}
\vspace{-1mm}

The spatial feature distribution of small objects in the COCO dataset is often very different from that of small objects in the real world. A trainable object filter was employed to bridge the gap between the two distributions. This filter is a binary classifier that classifies a given image into realistic or unrealistic object classes. 
We use the edge image as input for binary classifier. Binary classifiers are required to have transferability and to be able to capture synthetic imageness. As evidenced by existing research\cite{chen2020harmonizing}, low desctiminarity image has high transferability.  In addition, the edge image may emphasize the unnaturalness because the synthetic image has unnatural object boundaries.
The classifier was trained in two steps. First, a training image was cropped using the bounding box of the detected object region. Next, the cropped subimage was converted to an edge image using the Sobel filter operation and then resized to 224 $\times$224. The object edge images obtained in this manner were regarded as positive class training samples. Random images were used as negative class samples. Subsequently, a convolutional neural network (CNN) with ImageNet\cite{deng2009imagenet} pretrained weights was defined as a binary classifier. 

\subsection{Training Procedure}
\vspace{-1mm}

Ideally, the change detector should be adapted to a new domain while sharing parameters with the old domain without suffering from catastrophic forgetting. Therefore, slightly different training schemes were used for the CD encoder and decoder of the change detector. Specifically, a Siamese network with a convolutional encoder (CSCD-Net \cite{sakurada2020weakly}) was considered in the change-detector architecture. The parameter set of its encoder is set to be trainable only when training the first generation but frozen when retraining the second and subsequent generations.

\section{Experiments}
\subsection{Dataset}
The dataset was collected using a mobile robot equipped with a camera in various real-world indoor workspaces, including convenience stores, conference rooms, living rooms, and corridors. Small objects were placed manually in each workspace. Two independent groups were included in this study. The first group comprised smartphones, cables, notepads, and pens. The second group comprised handkerchiefs, wallets, and IC cards. In this way, a total of 8 types of data sets were created for each location and each group. Each dataset consists of 82 to 314 images.

The robot was a two-wheel-type differential robot with a size of 345mm $\times$ 335mm $\times$ 450mm. An onboard front-facing RealSense D455 camera was used to collect images with size 640 $\times$ 480. Mapping and navigation is performed 2D LiDAR SLAM algorithm from the onboard 2D LiDAR and IMU data. Point-goal navigation was used as the robot navigation strategy, in which the action plan of the robot was represented by a sequence of future subgoal points. The consecutive subgoals were then connected by straight paths. The same subgoal sequence was used for both training and testing (i.e., deployment).

The experimental procedure for the workspace was as follows: First, the robot navigated the workspace for the first time, and the collected image sequence was inserted into the knowledge base. Next, the knowledge base was initialized with nine types of COCO objects, and a change detector was trained using the knowledge base. The small objects in group1 were then manually and randomly placed in the workspace. The robot then performed a second round of navigation and change-detection tasks in the workspace. The knowledge base was then updated with the detected real-world small objects, and the change detector was retrained using the updated knowledge base. Subsequently, the small objects in group1 were manually removed from the workspace, and the small objects in group2 were randomly placed. The robot then performed a third round of navigation and change-detection tasks in the workspace. Using the procedure described above, a dataset triplet consisting of training, retraining, and testing sets was obtained for a specific workspace. An additional triplet was created by swapping the roles of group1 and group2.

\subsection{Settings}
A correlated Siamese change detection network (CSCD Net) \cite{sakurada2020weakly} was used as the backbone of the change detection model. The architecture was based on a Siamese network using Res-Net18 \cite{he2016deep} as the encoder/decoder. The numbers of iterations for training and retraining were set to 40,000 and 10,000, respectively.  The Adam algorithm, with a learning rate of 0.0001 is used and the batch size was 32. The number of training iterations for object prior filtering was 10 epochs. The SGD algorithm, with a learning rate of 0.001 is used and batch size was 32. The input live and reference images were pre-aligned using the state-of-the-art image alignment algorithm PDC Net \cite{truong2021learning}. All F-scores were calculated on aligned images. Nine groups of COCO objects consisting of apples, bottles, forks, spoons, toothbrushes, bananas, cups, handbags, and sports balls were used in the experiments.

\begin{figure}[t]

  \begin{center}
    \includegraphics[height=37mm]{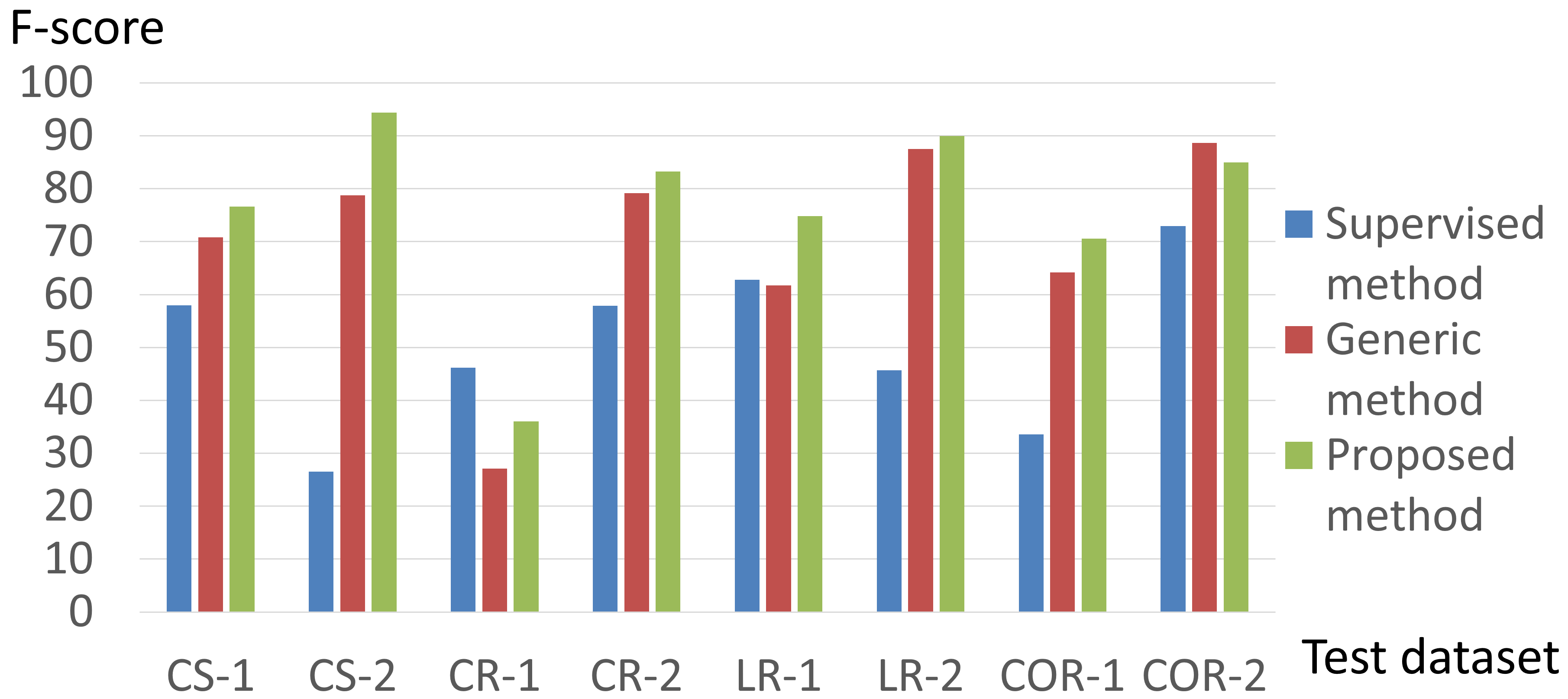}
  \end{center}
  \vspace{-4mm}

  \caption{
    Comparison with our proposed method and conventional method.
  }
  \label{kekka}
  % \vspace{-5mm}

\end{figure}

\begin{figure}[t]

  \begin{center}
    \includegraphics[width=80mm]{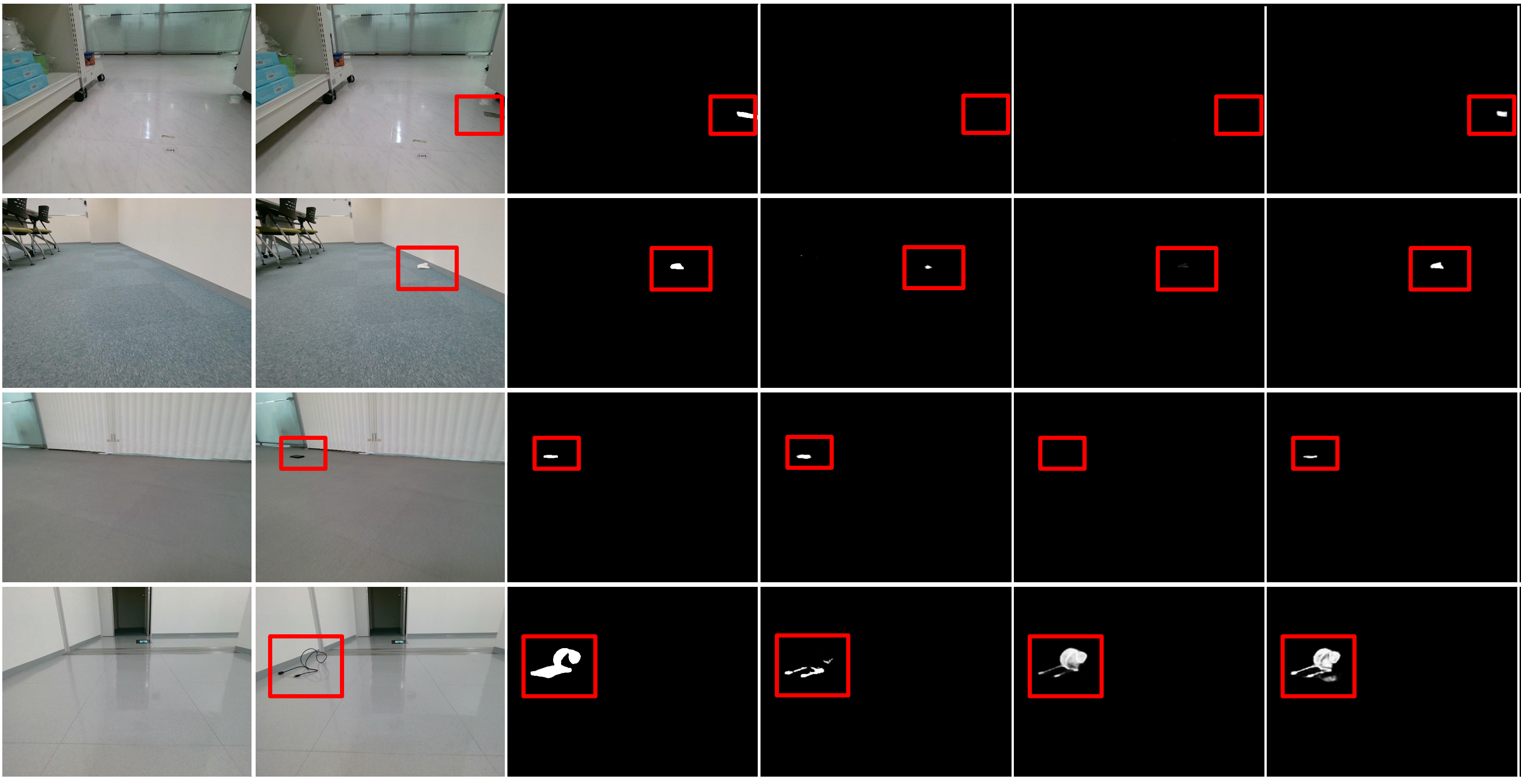}
  \end{center}
  \vspace{-4mm}

  \caption{
    Comparison with our proposed method and conventional method. From left to right, reference image, live image, ground truth, supervised method, generic method, proposed method.
  }
  \label{kekka_pic}
  \vspace{-5mm}

\end{figure}

\begin {table}[t]
  \caption {F-score of w and w/o object filter.}
  \label {object_filter}
\begin{tabular}{|l||r|r|r|} \hline
  Test Dataset & w/o filter & w filter & difference  \\ \hline
  CS-1 & 73.5 & 76.6 & {\bf +3.1} \\
  CS-2 & 79.0 & 94.4 & {\bf +15.4}\\
  CR-1 & 28.0 & 36.0  & {\bf +8.0} \\
  CR-2 & 87.4 & 83.2  & -4.2 \\
  LR-1 & 53.7 & 74.8  & {\bf +21.1} \\
  LR-2 & 86.7 & 90.0  & {\bf +3.3} \\
  COR-1 & 64.5 & 70.6  & {\bf +6.1}\\
  COR-2 & 88.7 & 85.0  & -3.7 \\ \hline
\end{tabular}
\vspace{-5mm}

\end{table}

\subsection{Results}
The performance of the change detector model was evaluated using the precision and recall of the pixelwise classification performance. The F-score, which is the harmonic mean of precision and recall, was also computed to evaluate the total performance.

% The performance of the change detector model was evaluated using F-score defined as follows,

% \begin{equation}
% F-score = 2\cdot \frac{Precision \cdot Recall}{Precision + Recall}
% \end{equation}

% \begin{equation}
%   Precision = \frac{TP}{TP + FP}
% \end{equation}

% \begin{equation}
%     Recall = \frac{TP}{TP + FN}
% \end{equation}

% where TP is the pixels where the prediction was change and the Ground Truth was also change, FP was the pixels where the prediction was a change and the Ground Truth was a unchange and FN is the pixels where the prediction was an unchange and the Ground Truth was a change.

Three different methods for training a change detector model, namely, a supervised method (supervised method), a generic object prior method (generic method), and the method proposed herein (proposed method), were evaluated and compared in terms of performance. The supervised method employs a CSCD-Net model trained in the test workspace and an independent object group; that is, group1 was used for training while group2 was used for testing, and vice versa. The generic method employs a CSCD-Net model trained on the test workspace and the aforementioned COCO small-object dataset. The proposed method, in which the model was initialized as the generic model, was deployed in the 1st generation navigation scenario and then retrained using the generic results of the 1st generation navigation and change detection.

Figure. \ref{kekka} summarizes the F-scores for the supervised, generic, and proposed methods CS refer to Convenience Store, CR refer to Conference Room, LR refer to Living Room, COR refer to Corridors, and the number at the end indicates the small object group. As shown in the figure, the proposed method outperformed the other two methods in most cases. Moreover, the generic method outperformed the supervised method in many cases. Figure. \ref{kekka_pic} shows the results of the three methods considered in this study.

By analyzing the strengths and weaknesses of each object in the proposed method, object group2 showed a better performance. This might be because continuous domain adaptation worked well because of the diversity of object shapes in object group1. On the other hand, the performance in the case of the conference room with the small object group2 was low. In this case, some objects were learned as part of the background and thus could not be recognized as foreground objects. The presence of a cable in the background image during training could have caused the network to learn the incorrect correspondence of the cable being the background. This may be resolved by increasing the number of generations of continuous domain adaptations.

Overall, a major error source was the lack of variety of foreground objects used when learning synthetic images, their divergence from real foreground objects, and the long-tail distribution of object features. 

These issues must be addressed in future studies.Table \ref{object_filter} shows the results with and without filtering object priors. The performance was higher with the filtering object priors.

% \begin{tabular}{|l||r|r|r|} \hline
%   Test Dataset & w/o filter & w filter & difference  \\ \hline
%   Convini-1 & 73.5 & {\bf 76.6} & + 3.1 \\
%   Convini-2 & 79.0 & {\bf 94.4} & + 15.4\\
%   Office-1 & 28.0 & {\bf 36.0}  & +8.0 \\
%   Office-2 & {\bf 87.4} & 83.2  & -4.2 \\
%   Living-1 & 53.7 & {\bf 74.8}  & +21.1 \\
%   Living-2 & 86.7 & {\bf 90.0}  & +3.3 \\
%   Hallway-1 & 64.5 & {\bf 70.6}  & +6.1\\
%   Hallway-2 & {\bf 88.7} & 85.0  & -3.7 \\ \hline
% \end{tabular}

\section{Conclusion}
In this paper, we lifelong change detection scheme for ground-view small object change detection. Our proposed method starts from domain knowledge containing noisy synthetic images and continuously performs domain adaptation through data collection and object prior filtering. As a result of experiments, it was confirmed that the performance greatly exceeds existing methods.

% \begin{thebibliography}{99}

% \bibitem{IMR}
%   I. M. Researcher, et al.: 
%   ``Read My Excellent Paper,'' 
%   \textit{Some Great Journal}, vol.xx, no.xx, pp.xx--xx, 200X.

% \bibitem{MVA}
%   MVA Conference: {\tt http://www.mva-org.jp/}

% {7795955,

%   author={Michael, Matthias and Feist, Christian and Schuller, Florian and Tschentscher, Marc},

%   booktitle={2016 IEEE 19th International Conference on Intelligent Transportation Systems (ITSC)}, 

%   title={Fast Change Detection for Camera-based Surveillance Systems}, 

%   year={2016},

%   volume={},

%   number={},

%   pages={2481-2486},

%   doi={10.1109/ITSC.2016.7795955}}

% \bibitem{MVA2}
% Anonymous, 202X.

% \end{thebibliography}
\bibliography{ref} %hoge.bibから拡張子を外した名前
\bibliographystyle{unsrt} %参考文献出力スタイル

\end{document}